\documentclass[sigconf,natbib=true,anonymous=false]{acmart}
\usepackage[ruled,linesnumbered]{algorithm2e}
\usepackage{multirow}
\usepackage{makecell}
\usepackage{enumitem}
\usepackage{diagbox}
\setitemize[1]{noitemsep,partopsep=0pt,parsep=0pt,topsep=0pt, leftmargin=10pt,
rightmargin=0pt}
\AtBeginDocument{%
  \providecommand\BibTeX{{%
    \normalfont B\kern-0.5em{\scshape i\kern-0.25em b}\kern-0.8em\TeX}}}

\setcopyright{acmcopyright}
\copyrightyear{2018}
\acmYear{2018}
\acmDOI{10.1145/1122445.1122456}

\acmConference[Woodstock '18]{Woodstock '18: ACM Symposium on Neural
  Gaze Detection}{June 03--05, 2018}{Woodstock, NY}
\acmBooktitle{Woodstock '18: ACM Symposium on Neural Gaze Detection,
  June 03--05, 2018, Woodstock, NY}
\acmPrice{15.00}
\acmISBN{978-1-4503-XXXX-X/18/06}

\begin{document}

\title{MATT-CTR: Unleashing a Model-Agnostic Test-Time Paradigm for CTR Prediction with Confidence-Guided Inference Paths}

\author{Moyu Zhang}
\affiliation{%
  \institution{Alibaba Group}
  \city{Beijing}
  \country{China}
}
\email{zhangmoyu@butp.cn}

\author{Yun Chen}
\affiliation{%
  \institution{Alibaba Group}
  \city{Beijing}
    \country{China}
}
\email{jinuo.cy@alibaba-inc.com}

\author{Yujun Jin}
\affiliation{%
  \institution{Alibaba Group}
  \city{Beijing}
  \country{China}
}
\email{jinyujun.jyj@alibaba-inc.com}

\author{Jinxin Hu}
\authornote{Corresponding Author}
\affiliation{%
  \institution{Alibaba Group}
  \city{Beijing}
  \country{China}
}
\email{jinxin.hjx@alibaba-inc.com}

\author{Yu Zhang}
\affiliation{%
  \institution{Alibaba Group}
  \city{Beijing}
  \country{China}
}
\email{daoji@alibaba-inc.com}

\author{Xiaoyi Zeng}
\affiliation{%
  \institution{Alibaba Group}
  \city{Beijing}
  \country{China}
}
\email{yuanhan@taobao.com}

\begin{abstract}
Click-through rate (CTR) models estimate the probability of a user clicking on an item by modeling the input features interactions. Recently, a growing body of CTR research has focused on either optimizing model architectures to better model feature interactions or refining training objectives to aid parameter learning, thereby achieving better predictive performance. However, previous efforts have primarily focused on the training phase, largely neglecting opportunities for optimization during the inference phase, where prediction performance is often degraded by low-frequency feature combinations. Although architectural components like attention or gating mechanisms aim to address this, they are susceptible to overfitting on these rare feature combinations and fail to mitigate noise during inference. Moreover, with the ever-growing complexity and parameter counts of models, training costs have escalated. Optimizing performance at test-time is therefore a crucial direction, as it not only improves prediction accuracy but also maximizes the utility of already-trained models, thus conserving training resources. Therefore, to address the above challenge, we propose a \textbf{M}odel-\textbf{A}gnostic \textbf{T}est-\textbf{T}ime paradigm (MATT), which leverages the confidence scores of feature combinations to guide the generation of multiple inference paths, thereby mitigating the influence of low-confidence features on the final prediction. Specifically, to quantify the confidence of feature combinations, we introduce a hierarchical probabilistic hashing method to estimate the occurrence frequencies of feature combinations at various orders, which serve as their corresponding confidence scores. Then, using the confidence scores as sampling probabilities, we generate multiple instance-specific inference paths through iterative sampling and subsequently aggregate the prediction scores from multiple paths to conduct robust predictions. Finally, extensive offline and online experiments strongly validate the effectiveness of MATT.
\end{abstract}

\keywords{Test-Time Paradigm,  Confidence-Guided Inference Paths, Click-Through Rate Prediction}

\maketitle

\section{Introduction}
\begin{figure*}[t]
  \centering
  \includegraphics[width=\linewidth]{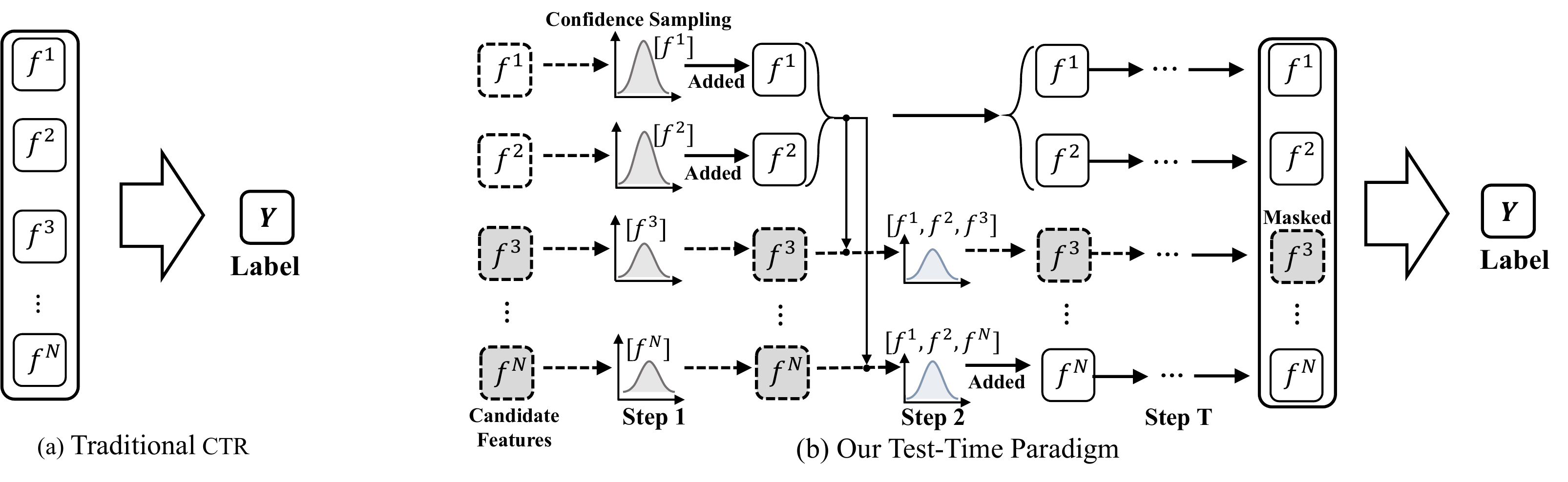}
  \caption{An illustration of previous methods and ours. Features in dashed boxes represent candidate features available for selection. (a) Previous approach utilizes the full set of input features for prediction. (b) Our method iteratively constructs a high-confidence inference path via probabilistic sampling, using the selected feature combination for scoring.}
  \label{example}
   \vspace{-0.3cm}
\end{figure*}

Click-through rate (CTR) prediction models are core modules of recommendation systems. They facilitate accurate recommendations by predicting a user's click probability on a target item \cite{back1, back2, back3, back4}, as shown in Figure \ref{example}(a). The prediction is derived from modeling the interactions among a vast combination of input features, encompassing user profiles and item attributes \cite{fm, deepfm, autoint, sfpnet, pepnet}. Therefore, effectively modeling these feature interactions is crucial for accurate prediction in the CTR prediction task \cite{genctr, dgenctr}. With recent advances in deep learning, numerous studies have proposed increasingly complex model architectures to capture correlations between features within input instances to boost prediction performance \cite{gate1, gate2, hstu}. Furthermore, drawing inspiration from scaling laws that have been proven to be applicable to most deep learning tasks, including Natural Language Processing (NLP) \cite{scal1} and Computer Vision (CV) \cite{scal2, scal3}, researchers have begun exploring generative paradigms to overcome the limitation of the discriminative paradigm, which are confined to a binary label space \cite{mtgr, dgenctr, onerec}.

In CTR research field, the conventional wisdom is that a model's inference performance is almost entirely determined by the effectiveness of its training. Consequently, while prior research has yielded significant performance gains by optimizing model architectures and training objectives \cite{dgenctr, mtgr}, these efforts have remained leaving the vast optimization potential of the inference phase largely unexplored. In practice, as established in prior work \cite{emb1, rcola}, while attention and gating mechanisms are designed for feature selection \cite{erase}, they often fail to provide reliable guidance for low-frequency combinations, as the model cannot learn to fit these sparse feature combination distributions adequately during training. As a result, the predictive power of a well-trained model can be severely compromised by these noisy features, preventing it from realizing its full potential. This challenge is further compounded by the trend towards ever-larger models in the CTR domain \cite{gen1, gen2, p5, tiger, lcrec, idrec}. The escalating parameter counts and associated training costs make frequent re-training or extensive hyperparameter tuning prohibitively expensive. Against this backdrop, exploring test-time optimization becomes not just beneficial, but essential. By strategically utilizing additional inference budget, a model can progressively explore more robust feature combinations for a given input sample, ultimately arriving at a more reliable prediction score, as illustrated in Figure \ref{example}(b). Such an approach not only enhances prediction accuracy but also maximizes the return on the substantial investment required for large-scale model training.

While recent works have begun to explore test-time paradigms in recommendation, these efforts have primarily focused on augmenting user sequences to enhance sequential recommendation performance \cite{tta}. Consequently, they fail to address the fundamental problem of feature combination sparsity that plagues CTR prediction. In the CTR task, low model confidence during inference is often a direct consequence of data sparsity, where numerous feature combinations appear too infrequently in the training set for the model to learn robust representations. This sparsity introduces high uncertainty, leading to unreliable predictions \cite{cl1, dgenctr}. For example, a feature combination that appears only a few times but is always associated with a positive label might mislead the model into assigning it strong predictive power—a classic case of overfitting. Theoretically, any feature combination with insufficient occurrences contains limited information, preventing the model from accurately learning its underlying distribution. This principle is well-established in traditional methods for mitigating the cold-start problem, which often replace a "low-confidence" item ID with more reliable, higher-level attributes like its category ID or associated popular multimodal features \cite{rd, lrd, scl}. By doing so, they substitute a sparse and unreliable signal with a robust one to ensure prediction stability. In this paper, we aim to develop a method, independent of the model's architecture, that helps selectively adjust the input features of an instance during inference. This approach is intended to compensate for the limitations of traditional gating or attention mechanisms in modeling sparse features or feature combinations, thereby significantly improving the confidence of prediction scores by leveraging the model's learned ability to identify reliable feature combinations.

Therefore, to address the issue of unreliable predictions caused by low-confidence feature combinations, which prevent a trained model from reaching its full potential, we propose the \textbf{M}odel-\textbf{A}gnostic \textbf{Test}-\textbf{T}ime paradigm (MATT) to explore multiple feature paths at inference time, thereby enabling instance-level adaptive input dynamic selection and leading to more accurate and robust predictions. MATT mitigates the impact of low-confidence feature combinations by quantifying their confidence scores and uses the scores as sampling probabilistic guides to generate diverse inference paths for the specific instance. By aggregating the predictions from these paths, MATT enhances the generalizability of the final output. Specifically, MATT is composed of two key modules: 

\textbf{Hierarchical Probabilistic Hashing}. Different from LLM-based inference methods which derive confidence score from token-level relationships \cite{ttl1, ttl2}, the binary objective of CTR models provides no equivalent mechanism for step-by-step verification. To develop a test-time paradigm applicable to most of CTR prediction models. we propose that use the occurrence frequency of features or feature combinations within the training set as a proxy for their confidence scores. The rationale is that a feature's frequency directly influences how reliably a model can learn its representation. However, the vast feature space of CTR models means that solely using hash tables to store these frequencies would lead to severe hash collisions, thereby compromising the accuracy of the confidence scores. By modeling the confidence of feature combinations as a probabilistic lower bound derived from hash-approximated occurrence frequencies, our hierarchical probabilistic hashing method accurately estimates the confidence of high-frequency features while simultaneously mitigating potential hash collision noise from less frequent ones.

\textbf{Confidence-Guided Paths Generation}. Once confidence scores for various feature combinations are obtained, our goal is to leverage these scores to construct more reliable final predictions. To this end, we introduce a novel multi-path generation method at inference time. Using the confidence of a new combination formed by a candidate feature and the current path as its sampling probability, we iteratively generate distinct feature selection paths for the specific instance, as shown in Figure \ref{example}(b). After multiple sampling steps, each path converges to a feature combination with a high confidence score. The feature combination from each path is then fed into the model to generate a prediction score, and these scores are subsequently aggregated to yield the final prediction based on each path's final confidence score. In this way, we mitigate the impact of unreliable features on the CTR model's output, enhancing its generalization and unlocking its full predictive potential, thereby improving overall prediction accuracy.

The contributions of our paper can be summarized as follows:
\begin{itemize}
\item To the best of our knowledge, MATT is the first work to introduce a model-agnostic test-time paradigm for the CTR prediction task, thereby mitigating the influence of low-confidence features and unlocking the predictive potential of the trained model.
\item MATT estimates the confidence of various feature combinations using the hierarchical probabilistic hashing method and uses the confidence scores as sampling probabilities to iteratively construct distinct inference paths during the inference phase. 
\item Evaluations using offline datasets and online A/B testing have demonstrated the compatibility and effectiveness of MATT with a wide range of existing state-of-the-art CTR models.
\end{itemize}  
 
\section{Related Work}
\subsection{CTR Prediction Models}
Deep learning has significantly advanced Click-Through Rate (CTR) prediction, leading to the widespread deployment of deep models in industrial recommendation platforms and garnering considerable research interest \cite{rel1}. At its core, CTR prediction seeks to model user click probabilities by capturing complex feature interactions, and consequently, a significant body of research has focused on designing sophisticated feature interaction architectures. Pioneering works include the Wide \& Deep Learning (WDL) model \cite{wdl}, which combined linear and deep components, and DeepFM \cite{deepfm}, which replaced the wide component with a Factorization Machine (FM) \cite{fm} for more effective low-order interactions. Others, like the Deep \& Cross Network (DCN) \cite{dcn, dcn2} and xDeepFM \cite{xdeepfm}, introduced explicit cross-networks and compressed interaction networks to model feature crosses. Subsequent work shifted towards using gating mechanisms to dynamically select salient features, such as EDCN's \cite{gate1} field-wise gating network and FRNet's \cite{gate2} complementary selection gate. In a different vein, RankMixer \cite{rankmixer} offers a scalable architecture by replacing self-attention with a token mixing module. Recently, recognizing that the binary objective in traditional discriminative CTR models can lead to suboptimal parameter learning, a separate body of work has shifted towards optimizing the model's objective by introducing generative paradigms. For instance, MTGR \cite{mtgr} preserves cross-features to prevent performance degradation, while DGenCTR \cite{dgenctr} reframes the task generatively using a discrete diffusion model. However, they overlook the significant optimization potential at inference time.

\subsection{Test-Time Scaling}
The performance of Large Language Models (LLMs) has seen remarkable progress in recent years \cite{llmd1, llmd2, llmd3, llmd4}. These improvements stem from two primary areas: advanced fine-tuning techniques, such as reinforcement learning frameworks that enable models to iteratively refine their outputs \cite{llmrl1, llmrl2, llmrl3}, and, critically, methods that exploit additional test-time computation \cite{ttl1, ttl2, ttl3, ttl4}. The latter has proven particularly effective, with studies showing that techniques like hierarchical hypothesis search can yield greater performance gains than simply scaling model parameters \cite{ttl1, ttl2}. This body of work establishes a clear trade-off, demonstrating the significant potential of inference-time optimization. However, while this work provides valuable insights, LLM-specific test-time methods are deeply coupled with their underlying architectures and task formats, rendering them non-transferable to CTR prediction models. This gap, combined with the proven potential of the test-time paradigm, underscores the urgent need for a model-agnostic test-time paradigm tailored specifically to the CTR prediction field. While TTA explored the test-time paradigm in recommendation, it focused on augmenting user sequences to improve sequential recommendation and fail to address the problem of feature combination sparsity that plagues CTR prediction \cite{tta}.

\section{Preliminary}
This section first defines the CTR prediction task and subsequently introduces the test-time process of the CTR models.

\subsection{CTR Prediction Task}

The objective of Click-Through Rate (CTR) prediction is to forecast the likelihood that a user will click on a specific presented item. The outcome of this prediction will determine the final ordering of items shown to users. A CTR prediction task is typically structured as a supervised binary classification problem.

\textbf{Problem Definition}  Given a complete set of features $\textbf{F}_{full}$ and the label space $y \in \left\{0, 1 \right\}$, the CTR prediction task aims to devise a unified ranking formula $\mathcal{F}: \textbf{F}_{full} \rightarrow  y$, to concurrently provide accurate, personalized prediction outcomes, indicating whether the target user will click the target item. The common feature space typically encompasses a variety of features that can be categorized into distinct fields, such as user-centric and item-centric features, which collectively represent an instance's comprehensive contextual information. Within this paper, the feature space $\textbf{F}_{full}$ is constructed as $[f^1, f^2, ..., f^N]$, where $N$ represents the number of the feature fields. Mathematically, the CTR prediction task involves estimating the probability that the target user will interact with the target item in a given contextual feature sets, as illustrated below:
\begin{gather}
P(y| \textbf{X}) =  \mathcal{F}(f^1, f^2, ..., f^{N})
\end{gather} 

\begin{figure*}[t]
  \centering
  \includegraphics[width=\linewidth]{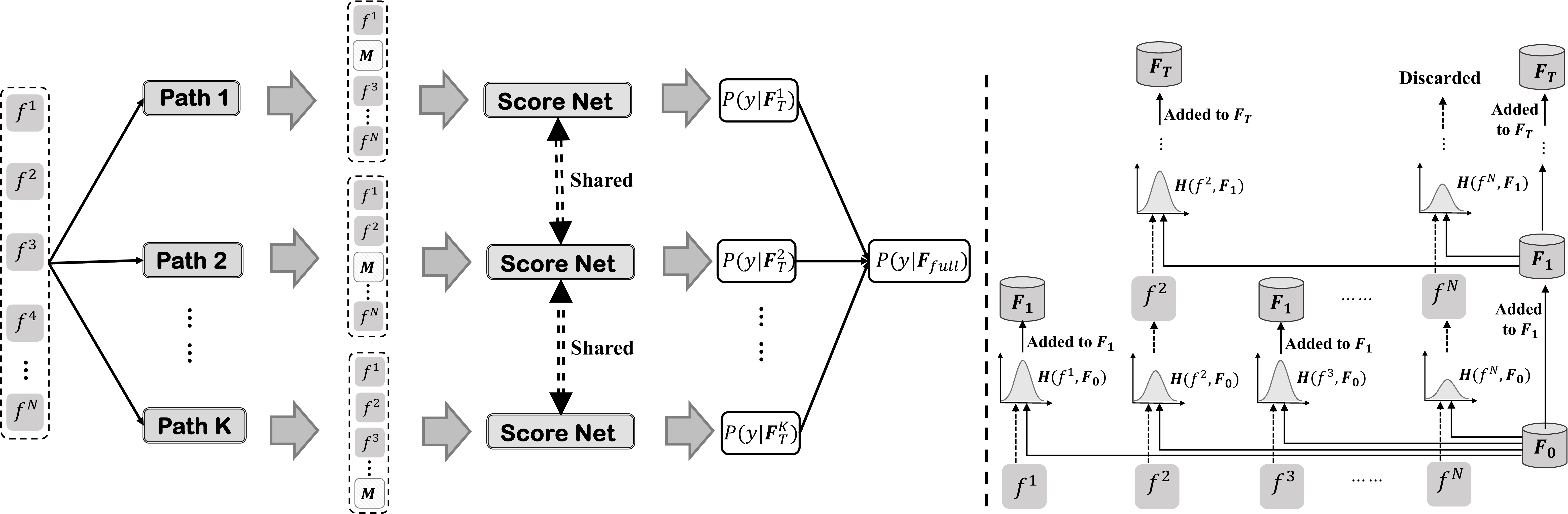}
  \caption{The structure of Model-Agnostic Test-Time framework (MATT) for CTR models. The left part shows the overall MATT scoring process, and the right part shows the specific inference path generation process.}
  \label{model}
   \vspace{-0.3cm}
\end{figure*}

\subsection{Test-Time in CTR}
Given a trained CTR model and an input instance consisting of $N$ features, conventional CTR methods directly feed the full feature set into the model to produce a score during inference phase. However, as previously discussed, data sparsity often leads to unreliable representations for infrequent feature combinations. Consequently, models can yield predictions with high uncertainty due to the representation collapse issue \cite{emb1, rcola, velf} and fail to unlock its full potential from training phase. Drawing inspiration from test-time scaling principles, additional computational resources can theoretically be leveraged during the inference process. Through step-by-step path exploration, it is possible to determine which features to use at each step, thereby identifying the most effective feature combination, that is, the optimal final inference configuration.

Formally, we can define $D(\theta, \textbf{F}_{full})$ as the distribution over output prediction outcomes induced by the model for a given complete feature set input $\textbf{F}_{full}$ for test-time computing based on the trained hyper-parameters $\theta$ of $\mathcal{F}(\cdot)$. We would like to select the hyper-parameters including model parameters and the features which maximize the accuracy of the target distribution for a given problem, which can be expressed as follows:
\begin{gather}
\mathcal{G}(\theta, \textbf{F}_{full}) =  \mathop{argmax}\limits_{\hat{y} \sim D(\theta, \textbf{F}_{full})}\left(\mathbb{E}[ \delta(\hat{y}=y)]\right)
\end{gather} 

\section{Method}
While prior CTR prediction research has enhanced performance through training-phase optimizations, such as architectural and paradigmatic changes \cite{din, pepnet, deepfm, mtgr, dgenctr}, the potential of test-time scaling has, to date, remained unexplored. Specifically, no prior work has sought to improve prediction performance by increasing computational effort at inference. This oversight is critical, as a persistent performance gap between the training and inference phase often remains for even a well-trained model, stemming from the data sparsity inherent in recommendation systems which leads to numerous low-confidence feature combinations \cite{emb1}. Although training-time regularization can alleviate overfitting on such data to improve generalization, the presence of these unreliable feature combinations during inference still compromises scoring accuracy and introduces significant prediction uncertainty. We contend that by adaptively identifying and pruning these unreliable feature combinations at inference time, while relying only on high-confidence features the model has learned well, it is possible to unlock the full predictive potential of a trained model and substantially improve its accuracy. Furthermore, the escalating training costs associated with the trend of ever-larger model parameters make efficient, test-time scaling an increasingly valuable pursuit. To this end, this paper pioneers a Model-Agnostic Test-Time paradigm (MATT) for the CTR task that generates sample-specific inference paths guided by feature combination confidence. This approach is composed of two key components, namely a hierarchical probabilistic hashing module and a confidence-guided paths generation module.\\
\textbf{$\bullet$ Hierarchical Probabilistic Hashing}: This stage quantifies the confidence of feature combinations, which is defined as their posterior occurrence frequency. It employs a hierarchical strategy: a min-heap is used to precisely track the counts of high-frequency feature combinations, while a probabilistic method estimates a lower-bound frequency for low-frequency feature combinations, thereby effectively mitigating the impact of hash collisions.\\ 
\textbf{$\bullet$ Confidence-Guided Paths Generation}: This stage iteratively samples candidate features into distinct inference paths. The sampling probability for each feature is determined by the confidence of the new combination it forms with the existing path. This ensures each path converges toward a feature set with high confidence score, which will be used to yield a robust final prediction.

\subsection{Hierarchical Probabilistic Hashing}  
Unlike inference methods for large language models \cite{ttl1, ttl2}, which can leverage intermediate training objectives to guide generation, CTR prediction models are typically trained on a singular and binary objective. This provides no direct mechanism for finding an appropriate optimization target during the inference phase. In fact, the challenge of improving performance in inference time is compounded by the inherent data sparsity in recommendation systems. When the model encounters rare or unseen feature combinations, its ability to model their interactions is compromised, often leading to representation collapse and uncertain predictions. Based on this observation, we posit that the confidence of a model's prediction is strongly correlated with the occurrence frequency of the underlying feature combination of the input in the training data. To operationalize this insight in a model-agnostic manner, we propose using the posterior occurrence count of a feature combination as its confidence score. The rationale is that if a feature combination appears frequently, the model will have learned to score it accurately. Conversely, if the combination is infrequent, any prediction based upon it will inherently carry high uncertainty.

However, the high dimensionality of features in CTR models makes it computationally infeasible to store exact occurrence counts for all possible feature combinations. To address this challenge, we propose a hierarchical probabilistic hashing method designed to accurately estimate the true counts by maintaining separate statistics for feature combinations of different frequencies. Specifically, for each order of feature combination, we employ multiple independent hash tables. The occurrence frequency of a given combination is then estimated by taking the minimum value from its corresponding entries across these tables. This strategy provides a tight upper-bound estimate of the true count, effectively mitigating the overestimation errors caused by hash collisions. The process is illustrated as follows:
\begin{gather}
\text{m-th order}: \textbf{H}(c_m)= hash(c_m) = \min_{\left\{h^m_{i}\right\}_{i=1}^{L_m}} \left( h^m_{i}(c_m)\right )
\end{gather} 
where $\textbf{H}(c_m)$ denotes the confident value of feature combination $c_m$, and $hash(c_m)$ is the upper bound of frequency of the $m$-th order feature combination $\left\{c_m \right\}$. $h^m_{i}(\cdot)$ denotes the $i$-th hash table of the $m$-th order feature interaction and $L_m$ denotes the number of hash tables of the $m$-th order feature interaction level.  

Recognizing that high-frequency feature combinations are critical for reliable model predictions, our goal is to record their occurrence counts with the highest possible fidelity. The above standard multi-hash method, however, cannot fully eliminate hash collisions, which compromises this accuracy. This limitation is particularly significant given the sparse distribution of recommendation data: a small number of high-frequency combinations coexist with a vast and long tail of low-frequency ones. This phenomenon motivates a shift from a uniform counting strategy to a differentiated and hybrid approach. Therefore, we propose to treat these two groups differently. For the crucial set of high-frequency feature combinations, we employ a min-heap to maintain their exact occurrence counts. For the remaining low-frequency combinations, we instead leverage the values from our multiple hash tables to model a probability distribution, from which we estimate the information lower bound to serve as their final confidence score.

To isolate high-frequency features from the effects of hash collisions, we implement a peeling strategy. For each order of feature combination, we maintain a dedicated set of a fixed size to store the most frequent feature combinations. We identify these top combinations by their maximum estimated frequency, which is determined by querying across the multiple hash tables. Once a combination is added to this high-frequency set, its estimated count is then subtracted from its corresponding entries in all hash tables. This process effectively purifies the hash tables, reducing count pollution and ensuring the remaining values more accurately reflect the frequencies of low-frequency combinations.
\begin{gather}
\textbf{Z}_m = \left\{ c_i, \, c_i \in Top_{|\textbf{Z}_m|}(\textbf{C}_m) \right\}
\end{gather} 
where $\textbf{C}_m$ denotes the set of all $m$-th order feature combinations, $\textbf{Z}_m$ is the min-heap of $m$-th order, and $|\textbf{Z}_m|$ represents its capacity.

For low-frequency feature combinations, the estimated count obtained by taking the minimum value across multiple hash tables serves as an upper bound on the true frequency. Due to hash collisions, the true count is expected to be lower, and the values retrieved from each individual hash table will exhibit variance. To estimate a more realistic value, we treat the set of counts obtained from the different hash tables for a single feature combination as a distribution. We then calculate the variance of this distribution to quantify the range of the true count of combination $c_i$:
\begin{gather}
\textbf{X}(c_i) = \left\{ h^m_j(c_i) \right\}_{j \sim [1, L_m], \, h^m_j(c_i) > 0}^n \\
\mu_{\textbf{X}}= \frac{1}{n} \sum_{h^m_j \sim \textbf{X}} (|h^m_j(c_i)|) \\
\sigma_{\textbf{X}}= \sqrt{\frac{1}{n} \sum_{h^m_j \sim \textbf{X}} (|h^m_j(c_i)|)^2 - (\frac{1}{n} \sum_{h^m_j \sim \textbf{X}}|h^m_j(c_i)|)^2}
\end{gather} 
where $\textbf{X}(c_i)$ denotes the set of $n$ hash values obtained for $c_i$. While the minimum of multiple hash table values provides an upper bound, we aim to derive a conservative lower bound from the distribution of $\textbf{X}(c_i)$. To do so, we use the distribution's mean $\mu_{\textbf{X}}$ and apply Chebyshev's inequality. This allows us to calculate the probability that the true count is less than a certain value relative to the mean, thereby establishing a probabilistic lower bound as follows:
\begin{gather}
P(\mu_{\textbf{X}} - x < k_1) \ge 1- \frac{\sigma_{\textbf{X}}^2}{k_1^2}, \, P(\mu_{\textbf{X}}- x > k_2) \le \frac{\sigma_{\textbf{X}}^2}{k_2^2} \\
P( x > \mu_{\textbf{X}} - k_1  | x < \mu_{\textbf{X}}-k_2) \ge \left(1- \frac{\sigma_{\textbf{X}}^2}{k_1^2} \right) /  \frac{\sigma_{\textbf{X}}^2}{k_2^2}
\end{gather} 
where we set $\mu_{\textbf{X}}-k_2=hash(c_i)$. If we let $k_1=1/\sqrt{\frac{1}{\sigma_{\textbf{X}}^2} - \frac{\alpha}{k_2^2}}$, then we get the lower bound probability higher than $1-\alpha$, i.e., $P( x > \mu_{\textbf{X}} -k_1) \ge 1-\alpha$ . In this way, the lower bound of $c_i$ can be $\mu_{\textbf{X}} - 1/\sqrt{\frac{1}{\sigma_{\textbf{X}}^2} - \frac{\alpha}{k_2^2}}$, which is denoted as $\textbf{H}(c_i) = \mu_{\textbf{X}} - 1/\sqrt{\frac{1}{\sigma_{\textbf{X}}^2} - \frac{\alpha}{k_2^2}}$. 

\subsection{Confidence-Guided Paths Generation}
Traditional CTR methods utilize the full set of input features for scoring at inference time. While some approaches attempt to mitigate the influence of low-frequency features through weighting schemes \cite{pepnet}, research indicates a more fundamental problem: the presence of such low-confidence features can cause representation collapse during feature interaction modeling, severely compromising prediction accuracy \cite{emb1}. Consequently, a trained model's predictive power can be undermined by the inclusion of even a single low-confidence feature. This problem highlights a key distinction between the objectives of training and inference. The training phase benefits from exposure to diverse, and even sparse, input patterns to learn robust feature interactions. In contrast, the inference phase should prioritize leveraging the combinations that the model has already learned to evaluate reliably, thereby avoiding the disruptive influence of low-confidence features. 

It is worth mentioning that since there is no absolute definition of whether a given confidence value of a feature combination is high or low, we prefer to explore multiple possibilities through random sampling to avoid the loss of effective feature information. Guided by this principle, we propose a novel inference paradigm that uses the confidence derived in the previous section. Our method iteratively constructs instance-specific feature paths by sampling features in proportion to their confidence scores. The final prediction is then based on these resulting high-confidence feature combinations, maximizing the utility of the trained model and improving estimation accuracy.

Specifically, the generation of an instance-specific inference path begins by initializing an empty feature set, denoted as $\boldsymbol{F}_0=\left\{ \varnothing \right\}$. In each subsequent iteration, a new feature is sampled from the set of unselected candidates of the model original and complete input feature sets $\boldsymbol{F}_{full}$. The sampling probability for a given candidate feature is determined by the confidence of the new combination it would form with the features already in the inference path, as defined by the following distribution:
\begin{gather}
p_i^t = \frac{\textbf{H}(f^i, \boldsymbol{F}_{t-1})}{\sum_{f^j \in \textbf{F}_{full}}^{f^j \not\in \boldsymbol{F}_{t-1}}\textbf{H}(f^j, \boldsymbol{F}_{t-1})} \\ 
\boldsymbol{F}_t = \boldsymbol{F}_{t-1} \cup \left\{ \mathbb{I}^t_i \right\}_{f^j \in \textbf{F}_{full}}^{f^j \not\in \boldsymbol{F}_{t-1}}
\end{gather} 
where $p_i^t$ denotes the normalized sampling probability of the feature $f^i$ over all unselected candidate features conditioned on the existing selected feature set $\boldsymbol{F}_{t-1}$ at the $t$-th step. $\textbf{H}(f^i, \boldsymbol{F}_{t-1})$ denotes the confident value of feature combination formed by adding $f^i$ to $\boldsymbol{F}_{t-1}$, as determined by our hierarchical hashing method.

For each candidate feature, we perform an independent Bernoulli sampling using its corresponding normalized sampling probability to decide whether to add it to the path. A successful trial results in the feature being added to the current feature set $\boldsymbol{F}_{t-1}$, where it becomes part of the conditioning context for the next step. An unsuccessful trial means the feature remains in the candidate set for future consideration. The process can be illustrated as follows:
\begin{gather}
\mathbb{I}^t_i = 
\begin{cases}
f^i, \,  r^t_i= 1 \\
\varnothing, \, r^t_i = 0
\end{cases}  \\
\boldsymbol{F}_t = \boldsymbol{F}_{t-1} \cup \left\{ \mathbb{I}^t_i \right\}_{f^j \in \textbf{F}_{full}}^{f^j \not\in \boldsymbol{F}_{t-1}}
\end{gather} 
where $r_i^t \sim Bernoulli(p_i^t)$ denotes the Bernoulli sampling on $f^i$.

Upon completion of $T$ iterations, this procedure yields a high-confidence instance-specific feature set $\boldsymbol{F}_T$. For the final inference phase, a new input is constructed by masking all features from the original input $\textbf{F}_{full}$ that were not selected for this path $\boldsymbol{F}_T$. This sparse input is then passed to the trained CTR prediction model to obtain the final prediction score, as follows:
\begin{gather}
\mathcal{G}(f^i) = 
\begin{cases}
f^i, \,  f^i \in \boldsymbol{F}_T \\
\textbf{0}, \, f^i \not \in \boldsymbol{F}_T
\end{cases}  \\
P(y|\boldsymbol{F}_T) = \mathcal{F}(\left\{ \mathcal{G}(f^i) \right\}_{f^i \in \textbf{F}_{full}} )
\end{gather} 

However, the path generation process described above is inherently stochastic, meaning a single sampled path is not guaranteed to find the globally optimal feature combination for the specific instance. For example, if a feature selected at an early step leads to low-confidence pairings with subsequent candidate features, this choice can trap the algorithm in a suboptimal path and prevent it from discovering better feature interactions. To mitigate this limitation, we extend our method to sample $K$ independent paths in parallel. This multi-path approach allows for a more comprehensive exploration of the feature space, capturing diverse and high-confidence interactions and producing a set of $K$ distinct final feature sets, that is, $L = \left\{ \boldsymbol{F}_T^1, \boldsymbol{F}_T^2, ..., \boldsymbol{F}_T^K\right\}$. 

The confidence for each path is determined by the aggregate confidence of its final feature combination. Given that the confidence levels for each path's feature combinations differ, we use these confidence levels as weights to aggregate the paths. This allows the prediction scores of high-confidence paths to have a greater influence, which aligns with our motivation to reduce the impact of low-confidence predictions. The final prediction is then calculated as a weighted average of the scores obtained from each path and help us capture feature information from multiple perspectives:
\begin{gather}
P(y|\textbf{F}_{full}) = \sum_{\boldsymbol{F}_T^i \in L}\frac{\boldsymbol{H}(\boldsymbol{F}_T^i )}{\sum_{\boldsymbol{F}_T^j \in L}\boldsymbol{H}(\boldsymbol{F}_T^j)}P(y|\boldsymbol{F}_T^i)
\end{gather}

\section{Conclusions}
While most research in CTR prediction has focused on optimizing model architectures and objectives during the training phase, the vast optimization potential of the inference phase remains largely unexplored. This is a critical oversight, as even well-trained models struggle to learn generalized representations for sparse features, often leading to low-confidence predictions and overfitting on spurious correlations. To address this gap and unlock the full predictive potential of trained models, we propose a Model-Agnostic Test-Time framework (MATT) specifically for CTR tasks, which leverages confidence of feature combinations to guide the generation of multiple inference paths, helping the model dynamically adjust its inputs. The framework consists of two core components: a hierarchical probabilistic hashing module and a confidence-guided paths generation module. Specifically, we calculate the posterior confidence of feature combinations at different frequencies to ensure the accuracy of the confidence of high-frequency feature combinations and use probabilistic estimation to model the confidence lower bound of low-frequency feature combinations. Based on the confidence of feature combinations, MATT iteratively samples features to generate feature inference paths, ultimately obtaining feature combinations with higher confidence, which helps estimate more robust scores. Finally, extensive experiments and online A/B testing validate the effectiveness and compatibility of MATT.

\end{document}